%% file: paper.tex
\documentclass[twocolumn]{fairmeta}
\usepackage{amsmath,amssymb}
\usepackage{dblfloatfix}
\usepackage{fontawesome5}
\setcitestyle{numbers,square,comma}

\title{Navigent: A Geometry-Grounded Agentic Framework for Vision Language Navigation}

\author[*]{Xinyu Zhou}
\author[*]{Weiren Cai}
\author[*]{Yitong Chen}
\author{Shenghai Yuan}
\author{Zihang Mu}
\author{Boyu Ma}
\author{Jiahua Dong}
\author{Yanshuo Lu}
\author{Jiaqi Bai}
\author[\dagger]{Jianfei Yang}

\affiliation{MARS Lab, Nanyang Technological University}

\contribution[*]{Equal contribution.}
\contribution[\dagger]{Corresponding author.\\[0.6em]
    \mbox{%
        \makebox[1.25em][c]{\color[HTML]{2962D8}\faGlobe}%
        \hspace{0.3em}%
        {\sffamily\bfseries Project Page:}%
    }\enspace
    \href{https://marslab-navigent.github.io/navigent.github.io/}{%
        \textcolor[HTML]{2962D8}{marslab-navigent.github.io/navigent.github.io/}%
    }\\[0.2em]
    \mbox{%
        \makebox[1.25em][c]{\color[HTML]{2962D8}\faGithub}%
        \hspace{0.3em}%
        {\sffamily\bfseries Code:}%
    }\enspace
    \href{https://github.com/MarsLab-Navigent/Navigent}{%
        \textcolor[HTML]{2962D8}{github.com/MarsLab-Navigent/Navigent}%
    }
}

\abstract{\input{sec/0_abstract}}

\correspondence{\email{xinyuzprf@gmail.com}, 
\email{jianfei.yang@ntu.edu.sg}}

\begin{document}
\raggedbottom
\maketitle

\input{sec/1_intro}
\input{sec/2_formatting}
\input{sec/3_Method}
\input{sec/4_Experiments}
\FloatBarrier

\par\medskip

\section*{LLM Usage Statement}
Our work focus on an agentic navigation framework.
GPT-6 is used for online navigation, visual route review,
and offline refinement of navigation prompts and GEC code,
without updating the model weights. GPT-6 also assisted
with manuscript language polishing, grammar correction,
and improving clarity and readability. No other generative
AI tools were used. The authors are responsible for all content.

\bibliographystyle{assets/ieeenat_fullname}
\setlength{\bibsep}{2pt}
\bibliography{paper}
\end{document}

%% file: sec/0_abstract.tex
Vision-and-Language Navigation (VLN) with general-purpose multimodal large language models (MLLMs) offers a flexible alternative to task-specific navigation policies, while reliable long-horizon execution remains challenging. Agents must track navigation progress, detect discrepancies between commanded and actual motion, and recognize when movement has passed relevant landmarks or entrances. We present Navigent, an agentic VLN framework built around a frozen MLLM and a Geometry-Grounded Execution Critic (GEC). GEC performs two complementary checks: action-to-motion consistency, which evaluates measured displacement against commanded movement, and motion-to-landmark consistency, which tracks agent-selected visual anchors and entrance candidates in world coordinates. Detected discrepancies interrupt the remaining action sequence and trigger reobservation, local adjustment, or node-based recovery. Historical visual memory supports progress tracking across extended routes. An offline self-improvement procedure further analyzes successful and failed training trajectories to refine navigation prompts and GEC logic while keeping model weights fixed. On the R2R-CE validation-unseen split (1,839 episodes)
and RxR-CE validation-unseen split(3,669 episodes),
Navigent achieves 84.0\%/86.0\% SR and 72.9\%/69.1\%
SPL, respectively, demonstrating the value of geometry-grounded execution verification for reliable agentic navigation.

%% file: sec/1_intro.tex
\section{Introduction}
\label{sec:intro}
\begin{figure*}[!t]
\centering
\includegraphics[trim=0 0 560bp 0,clip,width=\textwidth]{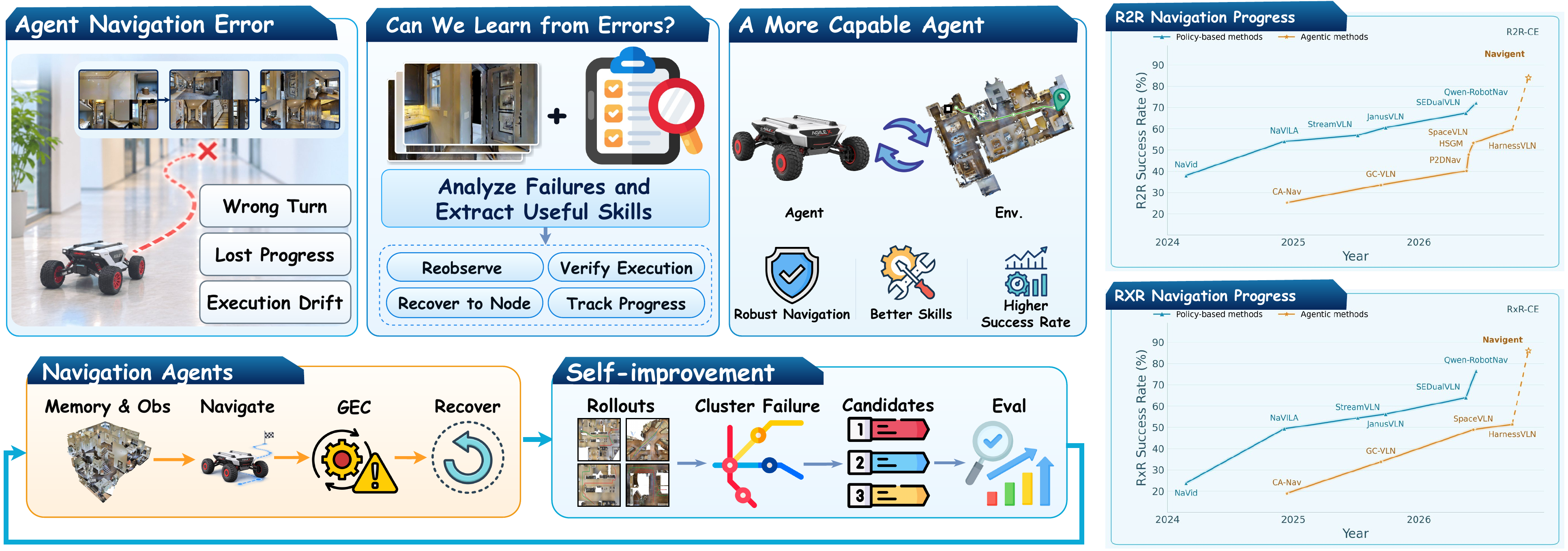}\par\vspace{3pt}
\includegraphics[trim=1306bp 326.08bp 0 0,clip,width=.49\textwidth]{figures/Motivation_navigent.compressed.pdf}\hfill
\includegraphics[trim=1306bp 0 0 326bp,clip,width=.49\textwidth]{figures/Motivation_navigent.compressed.pdf}
\caption{Navigent combines geometric execution checks with self-improvement. The plots compare navigation performance on R2R-CE and RxR-CE~\cite{navila,p2dnav}.}
\label{fig:motivation}
\end{figure*}
Vision-and-Language Navigation (VLN) enables embodied agents to navigate complex environments by following natural-language instructions and interpreting visual observations~\cite{r2r,vlnce}. Unlike conventional navigation toward predefined goals, VLN requires grounding linguistic descriptions of landmarks, spatial relations, and action sequences in the surrounding environment~\cite{r2r,rxr}. Reliable VLN remains challenging due to several factors. Long-horizon instructions require agents to continuously track navigation progress and relate current observations to past experiences~\cite{streamvln,mapgpt}. Visual ambiguity and occlusions complicate scene interpretation, while discrepancies between commanded and executed motions can cause route deviations~\cite{vlnce,harnessvln}. Local errors may also accumulate over extended navigation episodes, affecting both success rates and path efficiency~\cite{threestepnav}.

Existing VLN research has primarily followed two paradigms. Policy-based methods employ supervised learning, imitation learning, or reinforcement learning to train task-specific navigation policies that map visual and linguistic inputs to actions~\cite{etpnav,navid}. Recent navigation foundation models extend these policies across tasks and embodiments~\cite{navfom,lightnav,qwenrobotnav}, while dual-system models separate high-level planning from continuous action generation~\cite{dualvln}. In contrast, agentic VLN leverages the visual understanding, spatial reasoning, and tool-use capabilities of general-purpose multimodal large language models (MLLMs), enabling navigation through iterative reasoning and interaction with the environment~\cite{navgpt,agenticnav,supernav}. Recent navigation harnesses verify goals or planner proposals and use execution feedback to support recovery~\cite{navharness,harnessvln}. Translating high-level decisions into reliable continuous motion nevertheless requires monitoring what the robot actually executes. Consequently, achieving competitive success rates and path efficiency on full-scale benchmarks remains an important question for general-purpose agents.

We argue that reliable agentic navigation requires verifying both the execution of commanded actions and the spatial progress made relative to the environment. At the motion level, a forward command may produce insufficient displacement or unexpected lateral movement due to obstacles or execution deviations. At the landmark level, an agent may move forward while getting farther from a selected doorway, or pass an entrance without recognizing the resulting navigation error. These cases call for complementary geometric checks: comparing commanded actions with measured displacement, and tracking the robot's movement relative to visual landmarks. This motivates our central insight: navigation reliability can be improved by grounding execution verification in geometric evidence, alongside the agent's visual reasoning and action planning.

Motivated by this insight, we introduce Navigent, an agentic VLN framework built around a frozen MLLM. Historical visual memory integrates current observations with past visual evidence to support navigation-progress tracking. At its core, a Geometry-Grounded Execution Critic (GEC) performs action-to-motion verification using measured displacement and motion-to-landmark verification by tracking agent-selected visual anchors and entrance candidates in world coordinates. When GEC detects a discrepancy, it interrupts the remaining action sequence and returns control to the agent for reobservation, local adjustment, or node-based recovery. An offline self-improvement procedure analyzes successful and failed training trajectories to iteratively refine navigation prompts and GEC logic while keeping model weights fixed. On the complete R2R-CE (1,839 episodes) and RxR-CE (3669 episodes), Navigent achieves 84.0\%/86.0\% success rate (SR) and 72.9\%/69.1\% success weighted by path length (SPL)~\cite{spl}. 

\noindent Our contributions are summarized as follows:
\begin{itemize}
    \item {\interlinepenalty=10000 We develop a closed-loop agentic VLN framework around a frozen MLLM, integrating historical visual memory, incremental action execution, geometric verification, and node-based recovery to support reliable long-horizon navigation.\par}
    \item We introduce a Geometry-Grounded Execution Critic (GEC) that verifies action-to-motion and motion-to-landmark consistency through measured displacement, depth, camera geometry, and robot pose, enabling detection of execution deviations and possible landmark overshoots.
    \item We present an offline self-improvement procedure that analyzes successful and failed trajectories to refine navigation prompts and GEC logic through regression-aware candidate evaluation. On the R2R-CE and RxR-CE validation-unseen splits, Navigent achieves 84.0\%/86.0\% SR and 72.9\%/69.1\% SPL, respectively.
\end{itemize}

%% file: sec/2_formatting.tex
\section{Related Work}
\label{sec:related_work}

\subsection{Policy-based Vision Language Navigation}
VLN studies instruction following from visual observations, progressing from viewpoint graphs in R2R~\cite{r2r} to low-level control in continuous environments~\cite{vlnce}.
Learned navigation policies increasingly use MLLMs to encode visual history and predict actions~\cite{janusvln,correctnav,sedualvln}.
NaVid~\cite{navid} predicts actions from monocular video, while StreamVLN~\cite{streamvln} combines a sliding dialogue window with slowly updated visual memory to bound streaming context.
For broader task and embodiment coverage, NavFoM~\cite{navfom} unifies camera and temporal inputs, and LightNav-0~\cite{lightnav} connects spatial intent expressed through pointing to embodiment-specific trajectories through action tokenization.
Qwen-RobotNav~\cite{qwenrobotnav} exposes task modes and observation parameters so that a higher-level planner can configure and invoke the navigation model.
JanusVLN\cite{janusvln} decouples semantic and spatial
representations through dual implicit memory,
improving the modeling of instruction-relevant
visual context and navigation history across
continuous navigation tasks.

\subsection{Agentic Vision Language Navigation}
Agentic VLN organizes navigation as repeated reasoning and interaction through memory, skills, and execution tools~\cite{canav,gcvln,spacevln}.
NavGPT~\cite{navgpt} reasons over textual observations and history, while MapGPT~\cite{mapgpt} uses an online topological map to plan beyond local candidates.
AgentVLN~\cite{agentvln} connects high-level reasoning to perception and planning skills, including active depth queries and self-correction.
AgenticNav~\cite{agenticnav} exposes pixel-goal actions, depth queries, and visual recall as tools; SuperNav~\cite{supernav} combines navigation skills with interaction tools and task-progress management.

Execution verification extends these interfaces beyond action dispatch.
NavHarness~\cite{navharness} separates adaptive goal generation, goal verification, visuomotor execution, and persistent memory.
HarnessVLN~\cite{harnessvln} validates proposals against spatial evidence, geometric feasibility, and subgoal consistency before execution, then uses execution feedback to update memory and guide recovery.
Three-Step Nav~\cite{threestepnav} combines global planning and local grounding with retrospective trajectory review to correct accumulated drift before stopping.

\subsection{Self-Improvement}
Execution feedback can improve executable skills or model parameters~\cite{voyager,aspire,zetta,rsinav}.
Voyager~\cite{voyager} combines environmental feedback and self-verification to refine programs and accumulate skills.
ASPIRE~\cite{aspire} diagnoses failures from multimodal traces, validates program repairs through re-execution, and stores reusable robotic skills.
Zetta~\cite{zetta} keeps the base policy frozen while evolving runtime critics and recovery skills through rollout feedback and validation-gated updates.
Li et al.~\cite{rsinav} update navigation model weights by generating instructions for unlabeled trajectories, executing those instructions to filter samples, and retraining before processing the next trajectory shard.

%% file: sec/3_Method.tex
\section{Method}
\label{sec:method}

\begin{figure*}[!t]
\centering
\includegraphics[width=\textwidth]{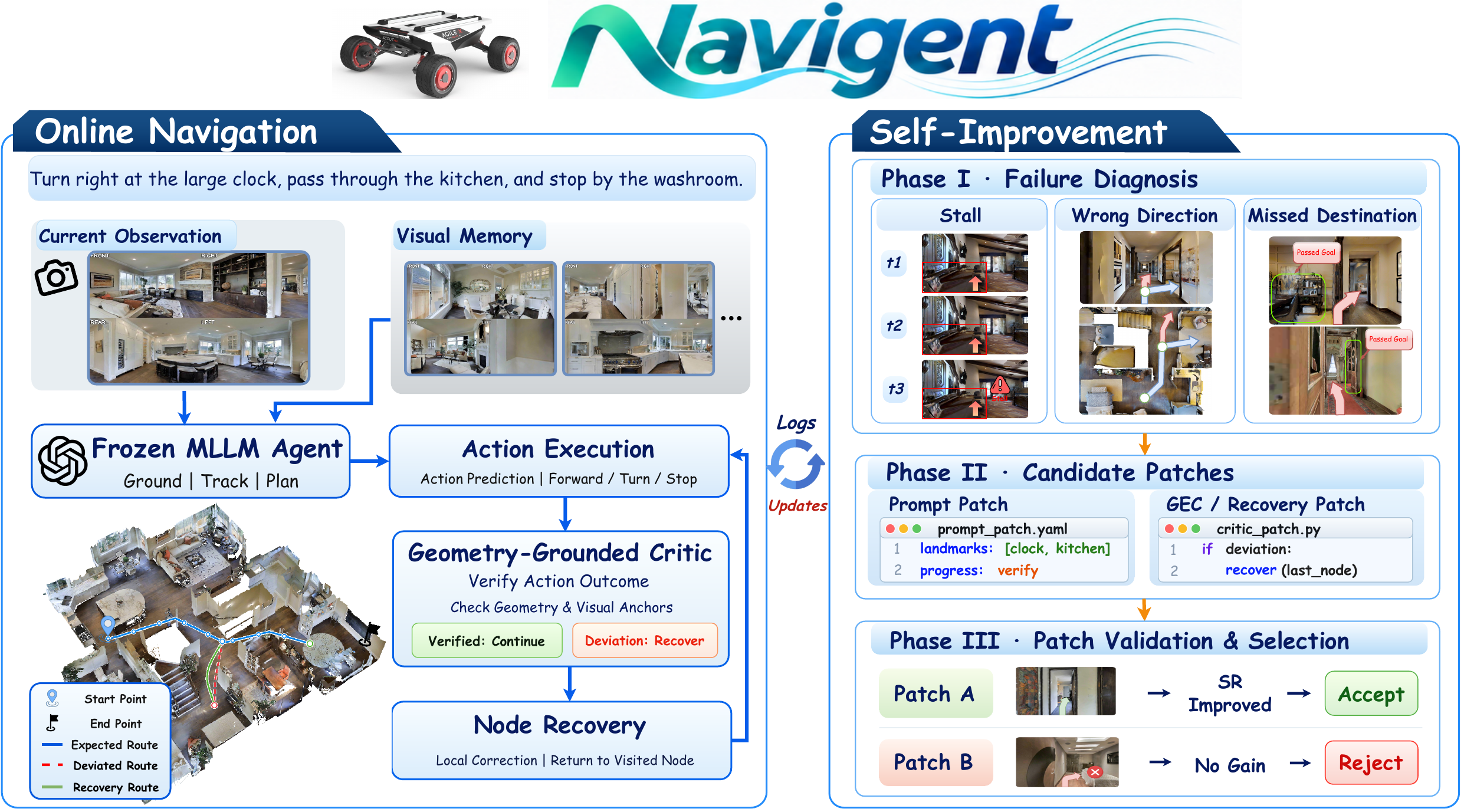}
\caption{Navigent uses visual memory and GEC for online navigation and refines prompts and critic code through offline self-improvement.}
\label{fig:overview}
\end{figure*}

Given a natural-language route instruction $I$, our agent uses successive visual observations to choose movement commands and decide when to stop at the described destination. As shown in Figure~\ref{fig:overview}, our method combines uniform historical visual memory with a Geometry-Grounded Execution Critic (GEC) and node recovery, while offline self-improvement refines navigation prompts and GEC logic. Visual memory supplies the current observation and up to eight images sampled across the executed route. The agent predicts a short action sequence, while GEC monitors measured execution and local visual anchors. A detected discrepancy interrupts the remaining sequence; the agent then reobserves and chooses a local adjustment or a return to a visited node. Completed training episodes provide evidence for refining navigation prompts and GEC implementation. The selected version remains fixed during navigation evaluation, while observation memory and execution state continue to update.

\subsection{Navigation Context and Execution}

\label{sec:memory}
\noindent\textbf{Historical Visual Memory.}
Following a route requires relating the current scene to previously encountered landmarks and branches. We retain a chronological observation archive and uniformly sample it at each decision, allowing the agent to inspect earlier route segments within a fixed budget of newly attached images.

\noindent\textbf{Observation representation.}
At decision time $t$, we form a composite image $P_t$ from directional RGB views, resized to a common display size, labeled by viewing direction, and arranged consistently. These labels associate landmarks with directions during movement reasoning. Current and historical observations share this representation; historical images retain their original orientations and separate camera perspectives. Camera configurations, resolutions, and layout are experimental settings.

\noindent\textbf{Archive and uniform selection.}
The archive starts empty. After each decision, we append its composite and frame identifier. Each frame represents that decision's multi-view observation; identifiers follow the cumulative primitive-step count and locate the image in the executed trajectory.

Let $n_t$ denote the number of archived observations before decision $t$, and $K_t=\min(n_t,8)$ the number selected. For $K_t\geq 2$, the selected archive indices are
\begin{equation}
i_j =
\left\lfloor
\frac{j(n_t-1)}{K_t-1}
\right\rfloor,
\quad j=0,\ldots,K_t-1.
\label{eq:history}
\end{equation}

For $K_t=1$, we select index zero; for $K_t=0$, the history input is empty. The rule retains every observation until the archive exceeds the budget, then samples across the full decision history, always including the earliest and most recent observations. Selection is based on chronological positions and requires no semantic keyframe ranking.

\noindent\textbf{Memory in decision making.}
At each decision, selected historical composites and frame identifiers accompany the current image $P_t$, which is outside the eight-frame history budget. The instruction $I$ enters the first decision; later decisions continue the same episode conversation. This context supports comparison of landmark identities, branch locations, and event order. The response records an estimated instruction stage, ordered event states, and visual evidence references as progress hypotheses. The eight-frame limit applies to newly attached historical images per decision; the conversation persists.

\noindent\textbf{Navigation State.}
We initialize each episode with the starting pose and empty trajectory, node records, and GEC state. During navigation, we record poses, visited nodes, and traversed connections. At each decision, the agent receives node identifiers and connectivity, recorded return-path lengths and relative return bearings, position relative to the start, horizontal heading, previous primitive counts, and the blocked flag. These records support ordinary navigation decisions and recovery after an interruption.

The executor uses self-pose and depth for geometric checks. Navigation uses recorded trajectories and node connections; dense RGB-D map construction and map-image input are disabled. Reference routes, goal coordinates, ground-truth room labels, and benchmark goal distances are excluded from navigation input.

\noindent\textbf{Action Prediction and Execution.}
The agent predicts discrete forward and turning commands, or selects STOP. To obtain new observations before completing the prediction, we execute a prefix of the proposed sequence. Let $H$ be the predicted horizon and $1\leq h<H$ the maximum number of selected commands:
\begin{equation}
\hat{\mathcal A}_t=(a_{t,1},\ldots,a_{t,H}),
\qquad
\mathcal A_t^{\mathrm{exec}}=(a_{t,1},\ldots,a_{t,h}).
\label{eq:prefix}
\end{equation}

The prefix includes forward motion and satisfies a total-travel bound. Commands expand into fixed-distance forward primitives and fixed-angle turns, indexed by $k$. Each primitive undergoes the checks in Section~\ref{sec:gec}, which can shorten execution. The agent then predicts from the updated observation. Invalid proposals receive bounded correction; action increments, horizons, and travel bounds are configurable.

\subsection{Geometry-Grounded Execution Critic and Node Recovery}
\label{sec:gec}

Historical images support route interpretation, while commanded actions may produce less motion or a different displacement than requested. We check measured execution after every primitive and interrupt the remaining sequence when a discrepancy is detected. GEC applies deterministic rules, refined offline through the self-improvement procedure in Section~\ref{sec:rsi} and fixed during each episode. The agent chooses the next movement or return node.

\noindent\textbf{Depth precheck.}
Before forward execution, the executor estimates near-field clearance from valid depths in a central image corridor. A near-obstacle detection interrupts the proposed movement before contact; insufficient depth yields an unknown result. This local check does not certify floor traversability.

\noindent\textbf{Measured motion.}
We compare each forward primitive with its resulting displacement. Let $\Delta p_k=(\Delta x_k,\Delta y_k,\Delta z_k)$ denote the measured displacement, with $y$ the vertical axis, and let $\theta_k$ denote the heading before execution, measured from the world $+x$ axis toward $+z$. We resolve horizontal displacement into forward and lateral components:
\begin{equation}
d_k^{\parallel}
=
\Delta x_k\cos\theta_k+
\Delta z_k\sin\theta_k,
\label{eq:forward}
\end{equation}
\begin{equation}
d_k^{\perp}
=
-\Delta x_k\sin\theta_k+
\Delta z_k\cos\theta_k.
\label{eq:lateral}
\end{equation}

Consecutive forward primitives with insufficient $d_k^{\parallel}$ trigger a stall interruption. A separate temporal check detects persistently small three-dimensional displacement. Once horizontal movement exceeds a minimum magnitude, negative forward progress or excessive lateral displacement triggers a direction conflict. These checks distinguish limited progress from movement inconsistent with the commanded direction.

For a declared stair maneuver, we evaluate signed vertical progress
$g_k=\sigma\Delta y_k$,
where $\sigma=1$ for ascent and $\sigma=-1$ for descent. Motion against this direction triggers an interruption. Consecutive forward primitives with negligible vertical progress also trigger reassessment of the approach or landing. All checks use configured displacement tolerances and persistence counts to test local execution consistency.

\noindent\textbf{Local visual anchors.}
To detect a possible overshoot of an agent-selected landmark, we complement motion measurements with a visual anchor. The agent selects a camera view and normalized image point for a visible target. The executor associates the point with a stable local depth estimate and projects it into world coordinates using camera geometry and measured pose. It then tracks the anchor's horizontal distance and bearing relative to the agent. An anchor represents a depth-supported surface point; a ray through a doorway may hit a surface behind the opening. Execution is interrupted when the anchor moves from ahead to behind the agent, or when forward motion increases its distance beyond a tolerance despite a predicted decrease.

The same geometric principle supports entrance monitoring. The agent proposes visible entrance candidates as normalized bounding boxes in named camera views, and the executor creates anchors at their centers. It maintains a bounded set of active candidates. When a nearby unresolved entrance moves behind the agent or its accumulated distance increases beyond the candidate threshold, the executor interrupts movement and carries an entrance-conflict alert into the next decision. GEC checks the geometry of these agent-selected locations; entrance identity, actual passage through an opening, and instruction interpretation remain agent judgments.

\noindent\textbf{Reobservation and node recovery.}
An interruption discards unexecuted primitives and returns control to the agent with a fresh observation, updated self-pose, primitive counts, and blocked status. Entrance-conflict alerts accompany the observation when present, and the complete GEC report is logged. The agent combines this evidence with historical images and node records to continue locally or revisit a node. Node access and backtracking remain navigation operations, while GEC determines when the current execution should end.

When the agent selects a return node, the executor replaces the predicted prefix in Eq.~\eqref{eq:prefix} with primitives following the recorded trajectory in reverse chronological order. Return execution has a bounded primitive count and uses the same execution checks and episode budget. Positional arrival is determined by a horizontal-distance tolerance around the selected node. An incomplete return carries the selected node and remaining distance into the next decision, where the agent inspects a new observation and decides how to proceed.

\noindent\textbf{Periodic route review.}
A periodic visual-review call examines the current observation and retained route keyframes. Its evidence-based assessment informs subsequent decisions; the navigator retains authority over movement and stopping. This semantic review complements the deterministic geometric checks.

\noindent\textbf{State update and termination.}
After execution, we record actual actions, update the trajectory and nodes, and archive the observation used for the decision. The next decision uses a new current observation and the history selected by Eq.~\eqref{eq:history}. The agent terminates navigation by predicting STOP with an empty movement list. The executor checks that the agent's ordered event list is nonempty and that all entries before its final entry are marked reached or passed. This check evaluates the consistency of the agent's event hypotheses; destination recognition remains its decision.

Episodes also terminate at the configured decision, primitive, or environment limit. Navigation prompts and GEC code are fixed within an episode, while runtime state is updated after execution. The backbone weights remain fixed throughout navigation and offline self-improvement. Benchmark goal distances are used only for evaluation.

\subsection{Self-Improvement}
\label{sec:rsi}

We use completed training episodes for offline self-improvement of navigation prompts and GEC implementation. Prompt updates refine the instructions guiding observation interpretation and action proposals; GEC updates refine movement checks and discrepancy handling. The MLLM weights remain fixed. Let $C_r$ denote the versioned agent at improvement round $r$, including its prompts and GEC code.

\noindent\textbf{Failure diagnosis from complete trajectories.}
We construct a fixed set
\[
D_{\mathrm{train}}=\{e_i\}_{i=1}^{N}
\]
by sampling episodes across training scenes and freeze it before the first round. This set is separate from the 100-episode Open-Nav ablation set. Each episode specifies an instruction, environment, and starting state. Running $C_r$ yields a binary success outcome $s_i(C_r)$ and records of observations, model responses, executed primitives, and measured motion. Failed episodes guide diagnosis; successful episodes serve as controls.

An offline GPT-6 author receives the current code, episode outcomes, a chronological decision index, and paths to complete records. It inspects inputs, outputs, and images on demand, groups failures by observable mechanism, and compares proposed actions with actual execution. Successful controls and competing explanations help identify recurring prompt or monitoring issues.

\noindent\textbf{Independent prompt and GEC updates.}
From this evidence, the author generates three candidate patches in one response, each applied to a separate copy of parent $C_r$. Each proposal targets a general navigation-prompt or execution-monitoring mechanism and includes supporting evidence, a competing explanation, an expected behavioral change, a risk to successful controls, and a condition that would falsify the diagnosis. All proposals are fixed before evaluation and applied to the same parent, so no candidate incorporates another candidate's evaluation results.

The author preserves the navigation model, observation inputs, historical-image budget, self-pose and visited-node access, action horizon, and scoring procedure. Updates target general navigation prompts and GEC runtime logic. Each patch undergoes applicability and syntax checks before evaluation.

\noindent\textbf{Evaluation of repairs and regressions.}
Each candidate is evaluated over the fixed training set, since a prompt or GEC change can repair failures and affect previously successful routes. We rerun every $C_r^{(j)}$ from each episode's original state with the same execution budget. Let $R_{r,j}$ count failure-to-success episodes and $L_{r,j}$ success-to-failure episodes. The selection score is
\begin{equation}
\begin{aligned}
\Delta_{r,j}
&=
\sum_{i=1}^{N}
\left[
s_i(C_r^{(j)})-s_i(C_r)
\right]\\
&=R_{r,j}-L_{r,j}.
\end{aligned}
\label{eq:netgain}
\end{equation}

This score measures the net change in successful episodes, so each regression offsets one repair. We also record SPL, decision and primitive counts, and model usage to characterize navigation efficiency and cost.

\noindent\textbf{Iterative version selection.}
Let $j^*$ index the candidate with the highest score in Eq.~\eqref{eq:netgain}. We update the parent by
\begin{equation}
C_{r+1}=
\begin{cases}
C_r^{(j^*)}, & \Delta_{r,j^*}>0,\\
C_r, & \text{otherwise}.
\end{cases}
\label{eq:selection}
\end{equation}

The selected patch alone is carried forward, while competing patches remain separate evaluation records. Its outcomes and trajectories, together with previous comparisons, inform the next round. If no candidate improves the score, the current parent is retained. We repeat this process for a fixed number of rounds and evaluate the selected prompt and GEC implementation. Each training rollout and evaluation episode uses a fixed version.

%% file: sec/4_Experiments.tex
\section{Experiments}
\label{sec:experiments}
\setcounter{dbltopnumber}{2}
\setlength{\textfloatsep}{8pt plus 2pt minus 2pt}
\setlength{\dbltextfloatsep}{12pt plus 2pt minus 2pt}
\setlength{\floatsep}{10pt plus 2pt minus 2pt}
\setlength{\dblfloatsep}{10pt plus 2pt minus 2pt}
\setlength{\intextsep}{10pt plus 2pt minus 2pt}

\input{tables/benchmark-comparison}

\begin{figure*}[!t]
\centering
\begin{minipage}[t]{\columnwidth}
  \centering
  \includegraphics[width=\linewidth]{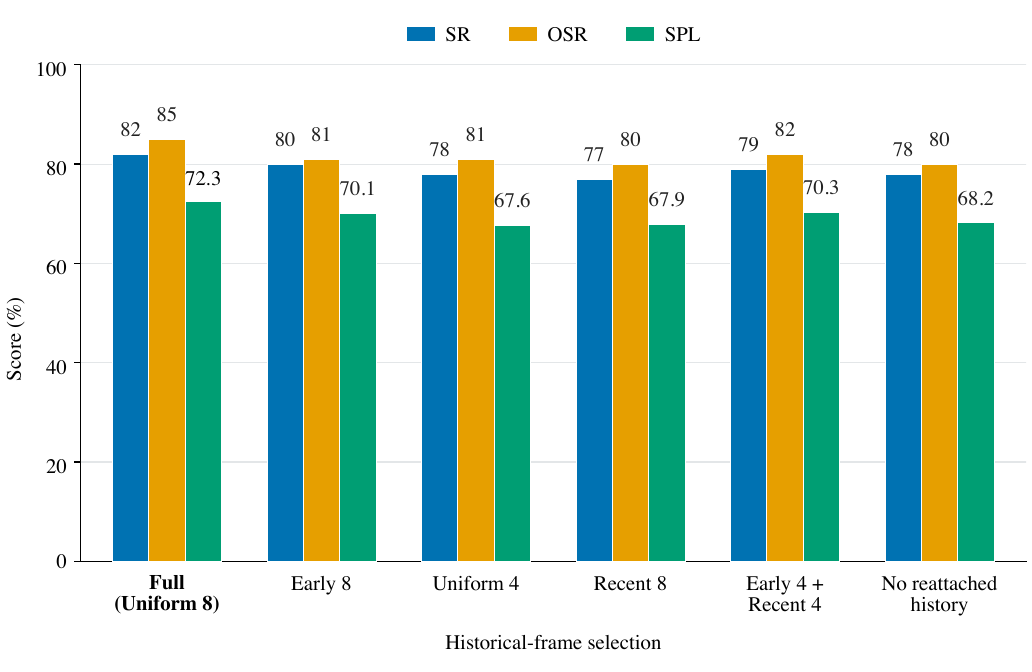}
  \caption{Navigent is evaluated with six historical-frame selection strategies on Open-Nav100.}
  \label{fig:history_ablation}
\end{minipage}\hfill
\begin{minipage}[t]{\columnwidth}
  \centering
  \includegraphics[width=\linewidth]{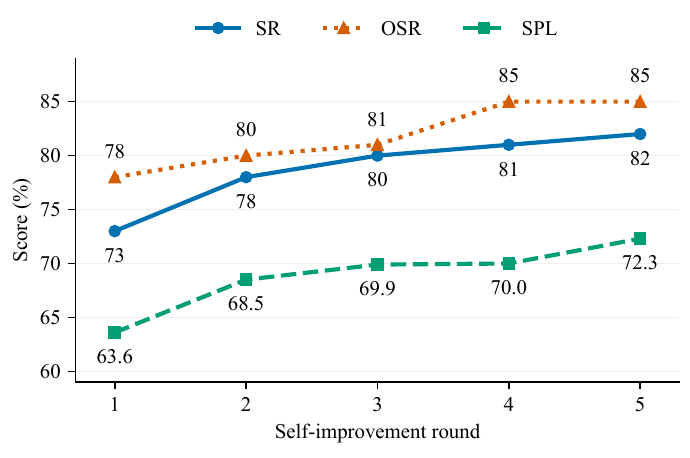}
  \caption{Navigent's SR, SPL, and OSR improve across five self-improvement rounds on Open-Nav100.}
  \label{fig:self_improvement}
\end{minipage}
\end{figure*}

\subsection{Experimental Setup}
\label{sec:experimental_setup}

\paragraph{Datasets and metrics.}
We evaluate Navigent in Habitat~\cite{habitat} on the validation-unseen splits of R2R-CE~\cite{r2r,vlnce} and RxR-CE~\cite{rxr}. Due to the API cost, ablations use the fixed Open-Nav100 subset~\cite{opennav}, separately from the episodes used to evaluate self-improvement candidates. We report navigation error (NE), success rate (SR), and success weighted by path length (SPL)~\cite{spl} on both benchmarks, plus oracle success rate (OSR) on R2R-CE and normalized dynamic time warping (nDTW)~\cite{ndtw} on RxR-CE.

\paragraph{Agent configuration.}
Navigent uses frozen GPT-6 with direction-labeled RGB observations, up to eight uniformly sampled historical frames, Habitat ground-truth self-pose, and visited-node records. GEC uses Habitat-rendered depth for geometric checks. The episode conversation persists across decisions. GEC monitors execution through motion checks, depth prechecks, and target anchors; periodic visual route review checks progress along the instructed route.

\subsection{Benchmark Results}
\label{sec:main_benchmark}

\paragraph{Main results.}
On R2R-CE validation-unseen (Table \ref{tab:benchmark_comparison}), Navigent achieves
84.0\% SR and 72.9\% SPL with frozen model weights.
HarnessVLN also builds its navigation agent on a general-
purpose MLLM. Compared with its reported full-split results, Navigent improves SR by 24.4 percentage points and
SPL by 29.3 points, while reducing NE from 4.08 m to 2.22
m. AgentVLN adopts an agentic formulation, while OmniNav employs a learned policy for navigation. Navigent improves SR and SPL by 16.8 and 8.2 points over AgentVLN,
and by 14.5 and 6.8 points over OmniNav. Compared with
OmniNav, NE decreases from 3.74 m to 2.22 m. On RxR-
CE, Navigent also achieves 86.0\% SR and 69.1\% SPL,
outperforming HarnessVLN, AgentVLN, and OmniNav by
34.6/31.0, 16.5/7.8, and 12.4/7.1 points in SR/SPL, respectively. It also reduces NE to 1.92 m, demonstrating consistent improvements across both benchmarks.


\subsection{Ablation Studies}
\label{sec:ablation_protocol}
\paragraph{Historical-frame selection.}
Figure~\ref{fig:history_ablation} compares six historical-frame selection strategies with the current observation, episode conversation, and execution critic fixed. No reattached history retains earlier dialogue; Early 4 + Recent 4 retains unique frames in temporal order.

Uniform 8 achieves the highest SR, OSR, and SPL. Compared with No reattached history, it raises SR from 78\% to 82\% and SPL from 68.2\% to 72.3\%. Other eight-frame strategies reach 77--80\% SR; Uniform 4 reaches 78\% SR and 67.6\% SPL.
\begin{figure*}[!t]
    \centering
    \includegraphics[width=\textwidth]
    {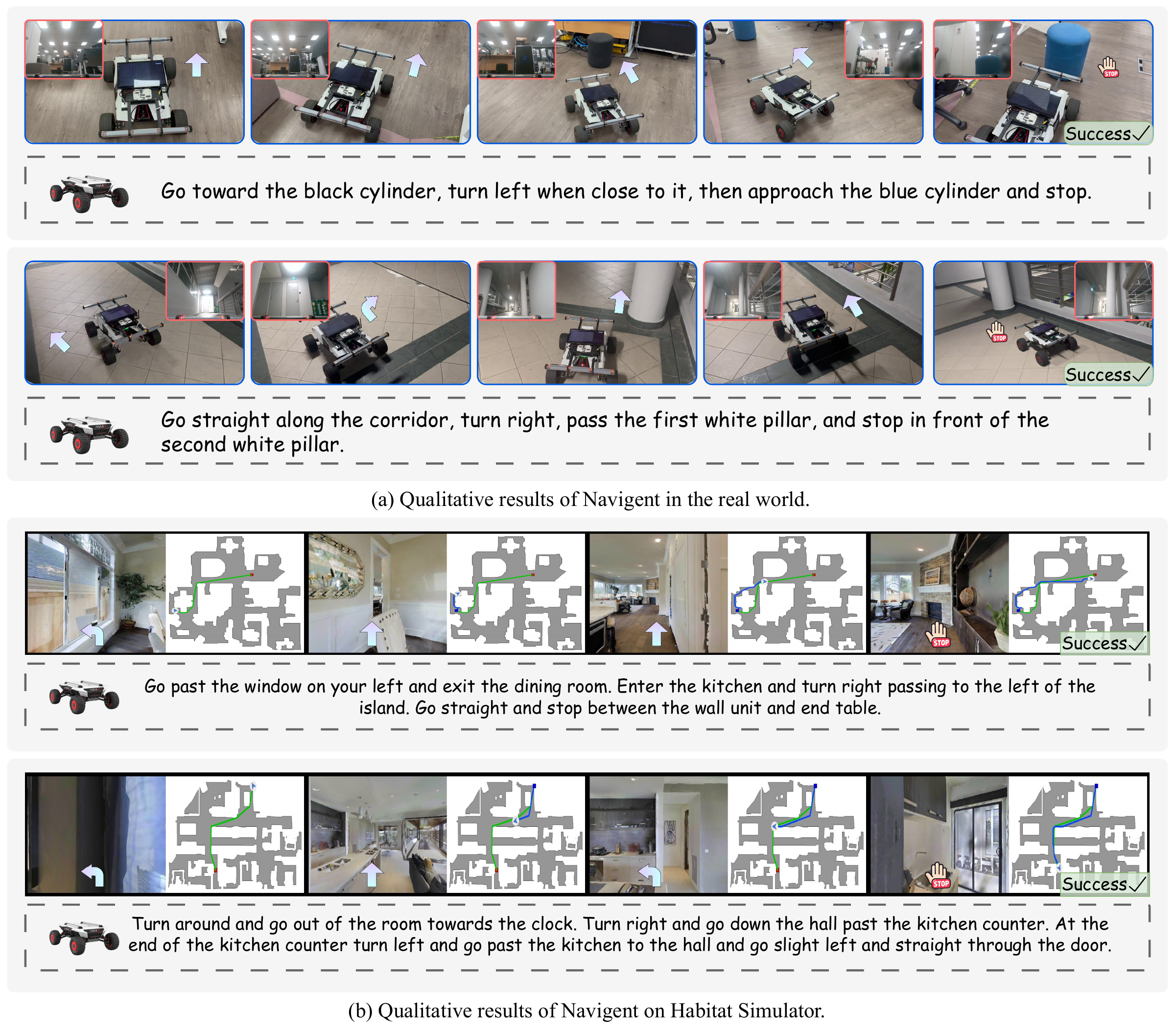}
    \caption{Navigent follows instructions and reaches the specified goals in real-world environments (a) and Habitat simulation (b).}
    \label{fig:qualitative}
\end{figure*}

\paragraph{Offline self-improvement.}
Figure~\ref{fig:self_improvement} evaluates five cumulative configurations on Open-Nav100 with frozen MLLM weights. The basic configuration uses measured pose changes to detect stalled forward motion. Prompt refinement adds explicit checks for landmark order, spatial relations, and visual evidence of progress, raising SR from 73\% to 78\%. Depth warnings check for obstacles before forward execution, complementing the motion checks performed afterward. Visual anchors track the distance and bearing of selected targets to detect movement inconsistent with the intended approach. Together, these additions raise SR to 81\% and OSR to 85\%. The final round adds periodic route review against the instruction and recorded observations, reaching 82\% SR, 72.3\% SPL, and 85\% OSR.

\input{tables/critic-ablation}

\par\addvspace{6pt}
\noindent\textbf{Execution critic.}
With the prompt fixed, the critic and route review
improve SR by 5 points and SPL by 2.8 points,
while reducing NE from 4.09 m to 3.54 m
(Table~\ref{tab:critic_ablation}). These gains suggest that execution
verification and route-level progress assessment
jointly improve completion and path efficiency
while reducing navigation error.

\par\addvspace{6pt}
\noindent\textbf{Self-pose and visited-node inputs.}
We ablate pose and node inputs while retaining executor geometry (Table~\ref{tab:state_session_ablation}). Removing either input reduces SR from 82\% to 79\%; removing both lowers it to 76\%. Without pose, NE improves despite lower SR and SPL.

\begin{figure}[!t]
    \centering
    \captionsetup{skip=3pt}
    \includegraphics[width=\linewidth,trim=0 10bp 0 11bp,clip]
    {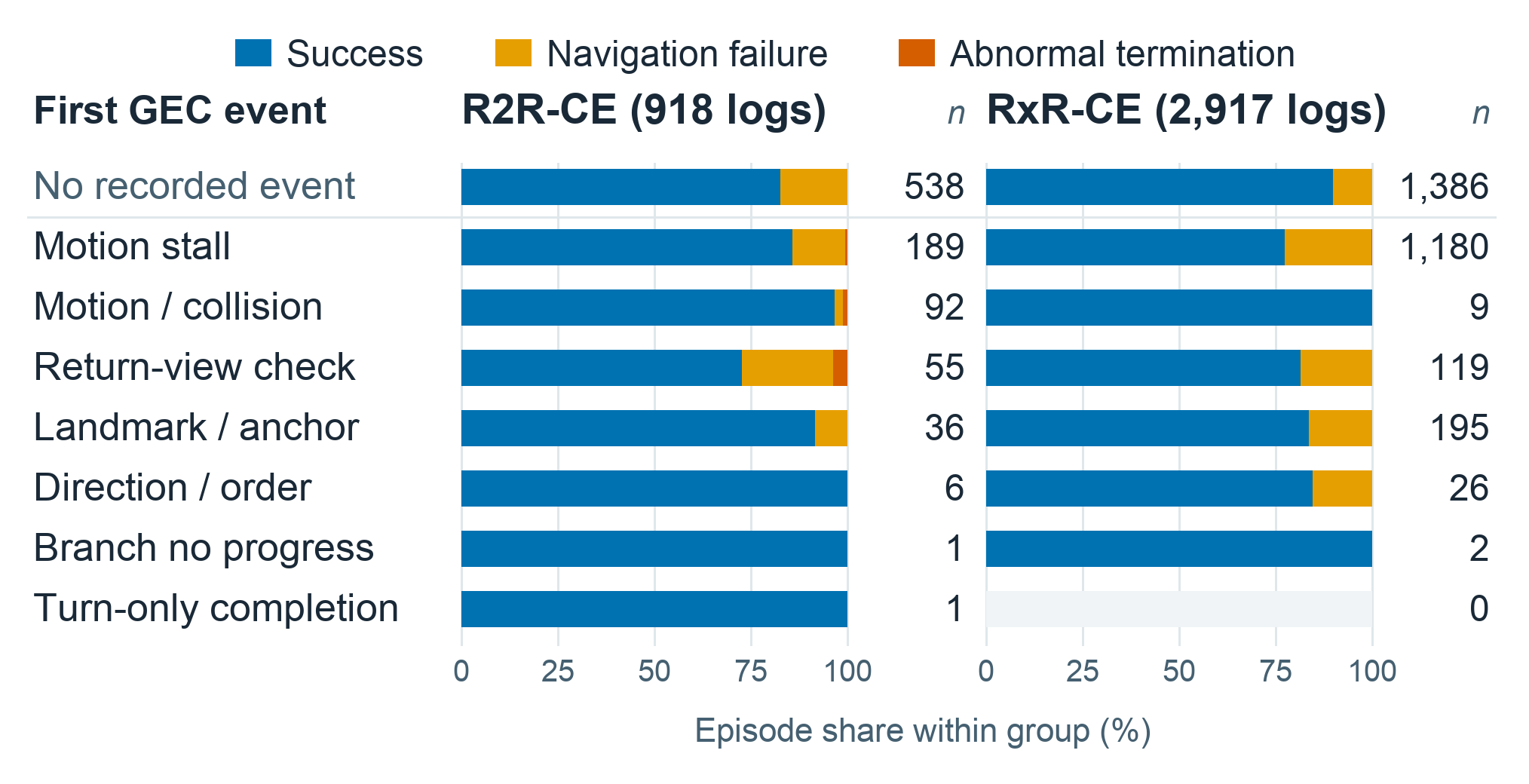}
    \caption{Episode outcomes by first recorded GEC event on logged R2R-CE and RxR-CE subsets.}
    \label{fig:gec_trigger_outcomes}
\end{figure}

\par\addvspace{6pt}
\noindent\textbf{Effects of different components.}
Table \ref{tab:rsi_ablation} examines the incremental contributions of each refinement component. Prompt refinement improves SR from 73\% to 78\%, while depth warnings further raise it to 80\%. Adding visual anchors increases SR to 81\% and OSR to 85\%, while reducing NE to 2.47 m. The final route review achieves the best SR and SPL of 82\% and 72.3\%, respectively, while maintaining 85\% OSR.

\par
\noindent
Figure~\ref{fig:gec_trigger_outcomes} groups the logged R2R-CE and RxR-CE simulation episodes by their first recorded GEC event. Episodes with no recorded event form a separate group. Within each group, the bars show the proportions of successes, navigation failures, and abnormal terminations.

{\captionsetup{skip=5pt}
\renewcommand{\arraystretch}{0.94}
\input{tables/state-session-ablation}



\input{tables/rsi-ablation}
}

\subsection{Real-World Evaluation}
\label{sec:real_world}

\input{tables/real-world}

\begin{figure}[!t]
    \centering
    \captionsetup{skip=3pt}
    \includegraphics[width=\linewidth]
    {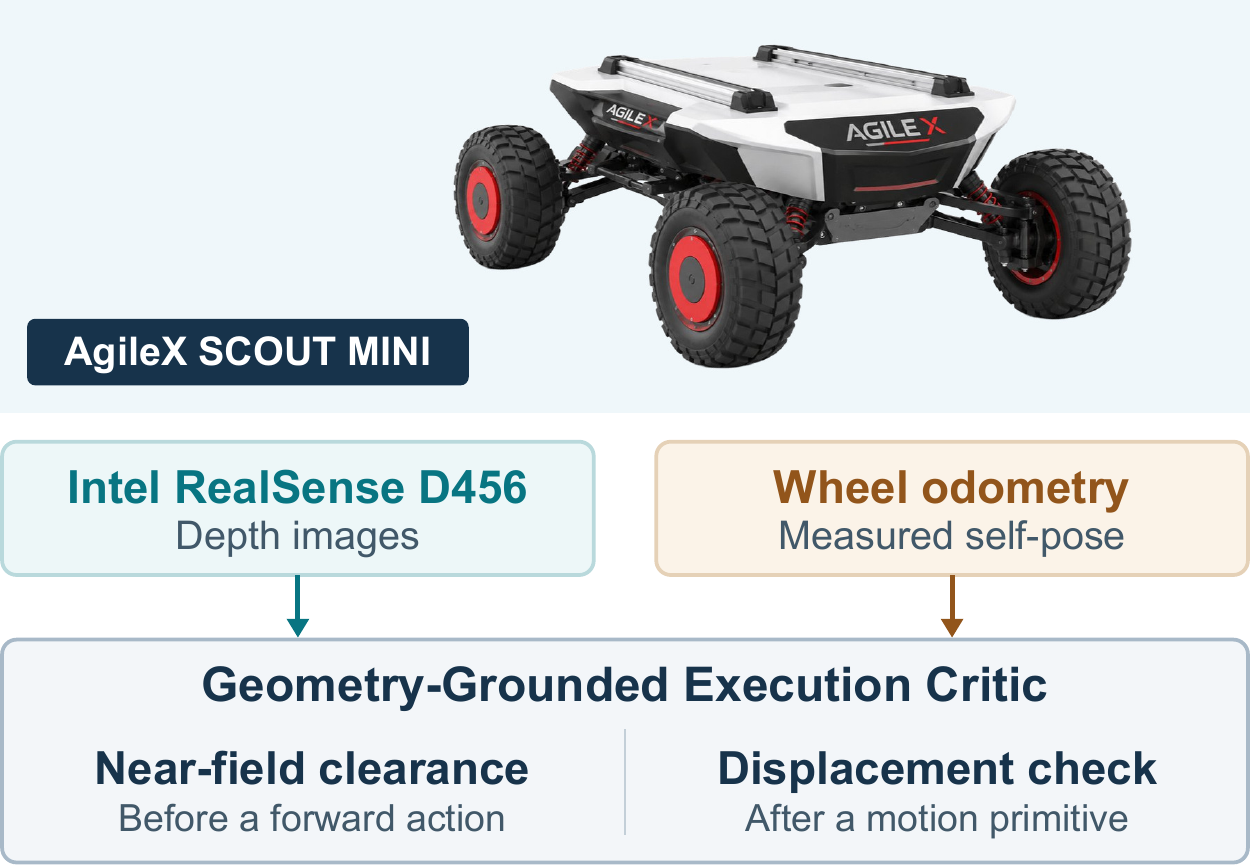}
    \caption{AgileX SCOUT MINI chassis and geometric inputs to the execution critic.}
    \label{fig:real_world_platform}
\end{figure}

We deploy Navigent on an AgileX SCOUT MINI with an Intel RealSense D456 RGB-D camera (Figure~\ref{fig:real_world_platform}). GEC uses wheel odometry and D456 depth, with calibrated camera-to-base extrinsics for target coordinate transformations. Each of two indoor and four outdoor tasks is evaluated five times. A trial succeeds when the robot executes STOP within 1\,m of the designated target. Navigent succeeds in 27 of 30 trials, achieving 90.0\% SR (Table~\ref{tab:real_world}). Figure~\ref{fig:qualitative} illustrates second-pillar identification.

\section{Conclusion and Limitation}
\label{sec:experiment_limitations}
We presented Navigent, a geometry-grounded agentic VLN framework that combines a Geometry-Grounded Execution Critic (GEC), and node-based recovery around a frozen MLLM. By verifying both action-to-motion and motion-to-landmark consistency, GEC detects execution deviations and potential landmark overshoots, enabling timely reobservation and recovery. Offline self-improvement further refines navigation prompts and critic logic using successful and failed trajectories without updating model weights. Currently, the self-improvement procedure focuses on refining navigation prompts and execution critic logic. Exploring more diverse forms of experience adaptation and knowledge reuse remains an interesting direction for future work.

%% file: tables/benchmark-comparison.tex
\begin{table*}[!t]
\centering
\caption{Comparison with sota methods on R2R-CE and RxR-CE val-unseen splits. \textemdash indicates the model was not evaluated on the metric.}
\label{tab:benchmark_comparison}
\begingroup
\setlength{\tabcolsep}{2.5pt}
\footnotesize
\begin{tabular*}{\textwidth}{@{\extracolsep{\fill}}llrrrrrrrr@{}}
\toprule
& & \multicolumn{4}{c}{R2R-CE Val-Unseen} & \multicolumn{4}{c}{RxR-CE Val-Unseen}\\
\cmidrule(lr){3-6}\cmidrule(lr){7-10}
Method & Source & NE$\downarrow$ & OSR$\uparrow$ & SR$\uparrow$ & SPL$\uparrow$
& NE$\downarrow$ & SR$\uparrow$ & SPL$\uparrow$ & nDTW$\uparrow$\\
\midrule
\multicolumn{10}{@{}l}{\textit{Policy-based methods}}\\
HNR~\cite{hnr} & CVPR~2024 & 4.42 & 67.0 & 61.0 & 51.0 & 5.50 & 56.3 & 46.7 & 63.5\\
NaVid~\cite{navid} & RSS~2024 & 5.47 & 49.1 & 37.4 & 35.9 & 8.41 & 23.8 & 21.2 & \textemdash\\
NaVILA~\cite{navila} & RSS~2025 & 5.22 & 62.5 & 54.0 & 49.0 & 6.77 & 49.3 & 44.0 & 58.8\\
Uni-NaVid~\cite{uninavid} & RSS~2025 & 5.58 & 53.5 & 47.0 & 42.7 & 6.24 & 48.7 & 40.9 & \textemdash\\
StreamVLN~\cite{streamvln} & ICRA~2026 & 4.90 & 63.6 & 56.4 & 50.2 & 5.65 & 54.4 & 45.4 & 63.7\\
JanusVLN~\cite{janusvln} & ICLR~2026 & 4.78 & 65.2 & 60.5 & 56.8 & 6.06 & 56.2 & 47.5 & 62.1\\
NavFoM~\cite{navfom} & ICLR~2026 & 4.61 & 72.1 & 61.7 & 55.3 & 4.74 & 64.4 & 56.2 & 65.8\\
CorrectNav~\cite{correctnav} & AAAI~2026 & 4.24 & 67.5 & 65.1 & 62.3 & 4.09 & 69.3 & 63.3 & 75.2\\
OmniNav~\cite{omninav} & ICLR~2026 & 3.74 & 74.6 & 69.5 & 66.1 & 3.77 & 73.6 & 62.0 & \textemdash\\
AwareVLN~\cite{awarevln} & CVPR~2026 & 4.02 & 73.5 & 65.4 & 55.1 & 3.95 & 67.6 & 56.1 & 65.7\\
\midrule
\multicolumn{10}{@{}l}{\textit{Agentic methods}}\\
InstructNav~\cite{instructnav} & CoRL~2024 & 6.90 & \textemdash & 31.0 & 24.0 & \textemdash & \textemdash & \textemdash & \textemdash\\
NavGPT-CE~\cite{navgpt} & AAAI~2024 & 8.40 & 26.9 & 16.3 & 10.2 & \textemdash & \textemdash & \textemdash & \textemdash\\
CA-Nav~\cite{canav} & TPAMI~2025 & 7.58 & 48.0 & 25.3 & 10.8 & 10.37 & 19.0 & 6.0 & 13.5\\
GC-VLN~\cite{gcvln} & CoRL~2025 & 7.30 & 41.8 & 33.6 & 16.3 & 8.80 & 33.8 & 13.8 & \textemdash\\
AO-Planner~\cite{aoplanner} & AAAI~2025 & 6.95 & 38.3 & 25.5 & 16.6 & 10.75 & 22.4 &15.1 &33.1\\
HSGM~\cite{hsgm} & CVPR~2026 & 5.42 & 58.7 & 47.9 & 32.8 & \textemdash & \textemdash & \textemdash & \textemdash\\
GTA~\cite{gta} & CoRL~2026 & 4.95 & 56.2 & 48.8 & 41.8 & \textemdash & \textemdash & \textemdash & \textemdash\\
HarnessVLN~\cite{harnessvln} & arXiv~2026 & 4.08 & 70.7 & 59.6 & 43.6 & 6.50 & 51.4 & 38.1 & 55.6\\
AgentVLN~\cite{agentvln} & ECCV~2026 & 3.88 & 73.5 & 67.2 & 64.7 & 3.92 & 69.5 & 61.3 & 74.6\\
\textbf{Navigent} & \textbf{Ours} & \textbf{2.22} & \textbf{86.1} & \textbf{84.0} & \textbf{72.9} & \textbf{1.92}& \textbf{86.0} & \textbf{69.1} & \textbf{79.7}\\
\bottomrule
\end{tabular*}
\endgroup
\end{table*}

%% file: tables/critic-ablation.tex
\par\addvspace{6pt}
\noindent\begin{minipage}{\linewidth}
\centering
\captionsetup{skip=4pt}
\captionof{table}{Effect of GEC and route review on Open-Nav100 with the navigation prompt fixed.}
\label{tab:critic_ablation}
\small
\setlength{\tabcolsep}{4pt}
\begin{tabular*}{\linewidth}{@{\extracolsep{\fill}}lrrrr@{}}
\toprule
GEC + review & SR$\uparrow$ & SPL$\uparrow$ & NE$\downarrow$ & OSR$\uparrow$\\
\midrule
Off & 77 & 69.5 & 4.09 & 80\\
Full & \textbf{82} & \textbf{72.3} & \textbf{3.54} & \textbf{85}\\
\bottomrule
\end{tabular*}
\end{minipage}
\par\addvspace{6pt}

%% file: tables/state-session-ablation.tex
\par\addvspace{10pt}
\noindent\begin{minipage}{\linewidth}
\centering
\captionof{table}{Effect of pose and visited-node inputs on Open-Nav100. The executor retains geometric feedback.}
\label{tab:state_session_ablation}
\small
\setlength{\tabcolsep}{2.5pt}
\begin{tabular*}{\linewidth}{@{\extracolsep{\fill}}lrrrr@{}}
\toprule
State inputs & SR$\uparrow$ & SPL$\uparrow$ & NE$\downarrow$ & OSR$\uparrow$\\
\midrule
Full & \textbf{82} & \textbf{72.3} & 3.54 & \textbf{85}\\
w/o Pose and Nodes & 76 & 70.6 & 3.32 & 83\\
w/o Nodes & 79 & 69.0 & 3.01 & 82\\
w/o Pose & 79 & 69.1 & \textbf{2.72} & 83\\
\bottomrule
\end{tabular*}
\end{minipage}
\par\addvspace{10pt}

%% file: tables/rsi-ablation.tex

\par\addvspace{6pt}
\noindent\begin{minipage}{\linewidth}
\centering
\captionof{table}{Effects of different components on Open-Nav100.}
\label{tab:rsi_ablation}
\small
\setlength{\tabcolsep}{2pt}
\begin{tabular*}{\linewidth}{@{\extracolsep{\fill}}lrrrr@{}}
\toprule
Method & SR$\uparrow$ & SPL$\uparrow$ & NE$\downarrow$ & OSR$\uparrow$\\
\midrule
Basic configuration & 73 & 63.6 & 3.10 & 78\\
+ Prompt refinement & 78 & 68.5 & 3.29 & 80\\
+ Depth warning & 80 & 69.9 & 3.05 & 81\\
+ Visual anchor & 81 & 70.0 &\textbf{2.47} & \textbf{85}\\
+ Route review (Full) & \textbf{82} & \textbf{72.3} & 3.54 & \textbf{85}\\
\bottomrule
\end{tabular*}
\end{minipage}
\par\addvspace{6pt}

%% file: tables/real-world.tex
\begin{table}[!htb]
\centering
\caption{Six real-world instructions with five trials per task.}
\label{tab:real_world}
\small
\setlength{\tabcolsep}{3pt}
\begin{tabular*}{\linewidth}{@{\extracolsep{\fill}}p{0.54\linewidth}rr@{}}
\toprule
Task & Successes & SR (\%)\\
\midrule
\multicolumn{3}{@{}l}{\textit{Indoor}}\\
Black, then blue cylinder & 4/5 & 80.0\\
Black--blue cylinder shuttle & 5/5 & 100.0\\
\midrule
\multicolumn{3}{@{}l}{\textit{Outdoor}}\\
Approach the gray chair & 5/5 & 100.0\\
Approach the green trash bin & 4/5 & 80.0\\
Stop at the second pillar & 4/5 & 80.0\\
Water dispenser, then green bin & 5/5 & 100.0\\
\midrule
\textbf{Overall} & \textbf{27/30} & \textbf{90.0}\\
\bottomrule
\end{tabular*}
\end{table}